# Possibilistic Assumption based Truth Maintenance System, Validation in a Data Fusion Application.

**Francesco Fulvio Monai and Thomas Chehire**

Thomson-CSF/RCC
160, Bd de Valmy,    BP 82
92704 Colombes Cedex. France
Tel: +33 (1) 47603985
Fax: +33 (1) 47603505
Email : Thomas.Chehire@eurokom.ie

## Abstract

Data fusion allows the elaboration and the evaluation of a situation synthesized from low level informations provided by different kinds of sensors. The fusion of the collected data will result in fewer and higher level informations more easily assessed by a human operator and that will assist him effectively in his decision process.
In this paper we present the suitability and the advantages of using a Possibilistic Assumption based Truth Maintenance System ($\Pi$-ATMS) in a data fusion military application.
We first describe the problem, the needed knowledge representation formalisms and problem solving paradigms. Then we remind the reader of the basic concepts of ATMSs, Possibilistic Logic and $\Pi$-ATMSs. Finally we detail the solution to the given data fusion problem and conclude with the results and comparison with a non-possibilistic solution.

## 1 A DATA FUSION APPLICATION

SEFIR (Système Expert de Fusion Interactive du Renseignement) is a prototype of a data fusion expert system which receives messages about observations provided by intelligence officers describing the nature, the number and the disposition of enemy units on the battle field and tries to derive the enemy formation [Lastic 89].
Units can be of several types (tank, motorised rifle, etc.) and are organised in hierarchical levels ranging from the higher division level down to the regiment, battalion, company and section levels. A data fusion process consists normally of three phases [Waltz 90]:

- the *correlation* phase, that is the association and combination of different informations concerning the same unit, obtained from different sources;

- the *aggregation* phase, that is the identification of a unit of a certain hierarchical level given partial evidence of its component units of the lower hierarchical level;

- the *fusion* phase, that is the elaboration of a(some) consistent situation(s) given partial informations provided by several sources.

The correlation phase is not implemented in SEFIR since we assume that each intelligence officer sends the informations concerning the units moving on the axis that has been allocated to him solely. Each unit is thus observed only once.

The messages received by SEFIR contain symbolic and numeric informations such as:

- the type and level of the observed unit (a tank section, for example);

- the time of the observation;

- the axis of enemy progress on which the observer is located;

SEFIR aggregates the observations described in the messages into higher hierarchical level units relying on its (incomplete) knowledge of the enemy's organisation and doctrine.

This knowledge is encoded in rules of the form:
"if three tank sections are observed on the same axis within an interval of time of one hour, then they may belong to the same tank company".

The uncertainty resulting from such incomplete domain knowledge induces the generation of a relevant number of somewhat contradictory hypotheses of possible aggregations of units for a given situation.

For example, if four tank sections are observed on the same axis within one hour, it is possible to create two hypotheses of aggregation at the superior hierarchical level of companies as shown in figure 1.

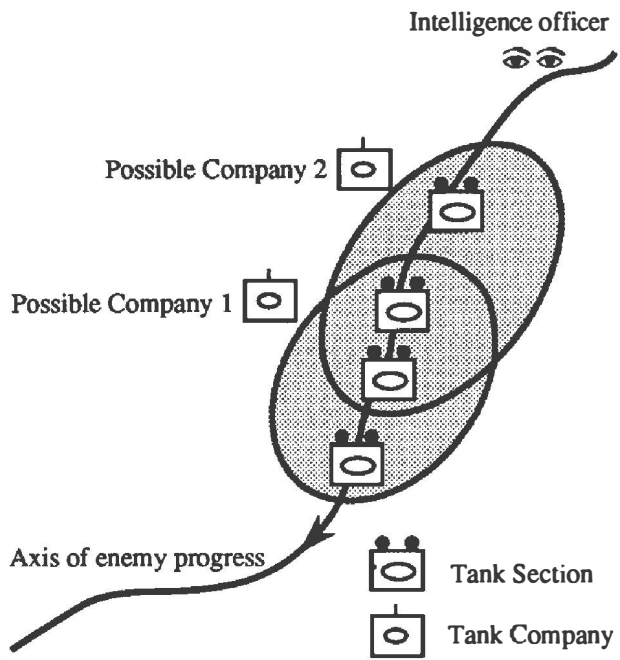

Figure 1: Creation of two contradictory hypotheses of companies, given evidence of four sections.

These two hypotheses are contradictory since they contain two common sections and thus they can't be both considered for a further aggregation at the battalion higher hierarchical level.

Moreover, these hypotheses are not equivalent since an expert would prefer the *Possible Company 1* hypothesis because it is more compact than the other one.

When dealing with such kind of uncertain knowledge one should be able to distinguish in the set of hypotheses the ones which are too uncertain to be considered as true, the ones which are almost certain and the ones which have an intermediary degree of certainty.

In particular, it is necessary to maintain such intermediary hypotheses to be able to use them in the reasoning process when the addition of new informations confirms their certainty.

An additional requirement comes from the fact that a data fusion system has to free the operator from low level details and to draw his attention on the higher level (fused) informations for a timely decision making process.

This can be achieved only if the possible contradictory choices provided by the system are limited enough. It is thus necessary to take advantage of all the available knowledge and induced constraints to limit the potential hypotheses to the most realistic (i.e. certain) ones.

In the case of SEFIR the knowledge which guides the discrimination among hypotheses is given by:

- the hierarchical structure of the observed enemy disposition (for example, a tank regiment contains three tank battalions and a motorised rifle battalion);
- the movement strategies of units depending on their level (for example, a regiment can move along two axis and all the units which compose it must be observed within two hours).

Three main features are thus requested for developing such a data fusion application:

- handling both numeric and symbolic data;
- handling multiple uncertain hypotheses somewhat contradictory;
- reducing these multiple hypotheses.

The first SEFIR prototype relied completely on the knowledge engineer for all the above stated functionalities. This resulted in adhoc algorithms that are neither proven to be correct nor easily reusable in similar applications by another knowledge engineer.

This paper describes a new prototype of the SEFIR expert system, using a Possibilistic Assumption based Truth Maintenance System ($\Pi$-ATMS) [Dubois 90], with proven logical foundations and that can ease the development of data fusion applications and the reuse of knowledge sources throughout many applications.

The suitability of a $\Pi$-ATMS arose from the ATMS support for exploring in parallel multiple contradictory hypotheses, and the Possibilistic Logic support [Dubois 88] [Lang 91] for formalizing imperfect knowledge, that is to express the uncertainty in a relative manner (one believes in an hypothesis more than in another).

## 2 ATMS BACKGROUND

ATMSs [deKleer 86] are Reasoning Maintenance Systems which make the distinction between *assumptions* (or *hypotheses*) and other data (or *facts*). Assumptions are data which are presumed to be true, unless there is evidence of the contrary. Other data are primitive data always true, or that can be derived from other data or assumptions.

Derivations of new data are recorded through *justifications* that link the newly created data to the assumptions or data that enabled their creation.

The ATMS is then in charge of determining which combinations of choices (assumptions) are consistent, and which conclusions they enable to draw.

## 3 POSSIBILISTIC LOGIC

Possibilistic logic is an extension of classical logic where ground formulas are weighted by two numbers belonging to the [0,1] interval, representing lower bounds of necessity and possibility degrees.

The possibility degree $\Pi(p)$ evaluates to which degree a proposition p is possible, that is coherent with the available knowledge.
The necessity degree $N(p)$ evaluates to which degree a proposition p is certain, that is implied by the available knowledge.

The following table summarizes the meaning of the weights attached to a proposition p:

| $\Pi(p)$ \ $N(p)$ | $N(p)=0$ | $0<N(p)<1$ | $N(p)=1$ |
|---|---|---|---|
| $\Pi(p)=0$ | p is false | Contradiction | |
| $0<\Pi(p)<1$ | p is somewhat false | | |
| $\Pi(p)=1$ | Ignorance | p is somewhat true | p is true |

However, given the application requirements we will restrict to lower bounds of a necessity measure $N$, provided by the application developer.
We can illustrate the updating of lower bounds of N, with a simple example that will be refined in the case of a $\Pi$-ATMS:
Let us have three propositions A, B, and C such that $N(A) \geq \alpha_A$, $N(B) \geq \alpha_B$, and $N(C) \geq \alpha_C$, and the justification $A \wedge B \rightarrow C$ such that $N(A \wedge B \rightarrow C) \geq \alpha_J$ ($\alpha_J$ denotes to which degree it is sufficient to believe in the premises A and B in order to believe in the conclusion C); then we can update N(C) with: $N(C) \geq \max(\alpha C, \min(\alpha_A, \alpha_B, \alpha_J))$.

We note that when simultaneously $0<N(p) \leq 1$ and $0 \leq \Pi(p)<1$ we have a (partial) contradictory situation since this means that both p and $\neg p$ are somewhat true.
Being able to pursue the reasoning in presence of partial inconsistency is one of the main issue of a Possibilistic ATMS.

## 4 POSSIBILISTIC ATMS

Possibilistic ATMSs are an extension of ATMSs to handle uncertainty in the framework of Possibilistic Logic.
In a $\Pi$-ATMS, hypotheses, facts and justifications can be uncertain. This allows to obtain a weight for the generated environments (i.e. a set of hypotheses supporting a datum), to evaluate the inconsistency degree of a set of hypotheses and to calculate the uncertain consequences of a set of uncertain hypotheses.

In a $\Pi$-ATMS the definitions of environment, nogood, label and context of a standard ATMS must be modified to take into account the uncertainty of hypotheses, facts and justifications.

Let J be a set of weighted justifications, H a set of weighted hypotheses, E a subset of H and d a datum.
Then we have the following definitions:

**Environments:** $[E \ \alpha]$ is an *environment* of d iff d can be deduced from $J \cup E$ with a certainty degree $\alpha$.
$[E \ \alpha]$ is an $\alpha$-*environment* of d iff $[E \ \alpha]$ is an environment of d and $\forall \ \alpha'>\alpha$, $[E \ \alpha']$ is not an environment of d ($\alpha$ is maximal).

**Nogoods:** $[E \ \alpha]$ is an $\alpha$-*contradictory environment*, or $\alpha$-*nogood* iff $J \cup E$ is $\alpha$-*inconsistent*, that is $\bot$ can be deduced from $J \cup E$ with $\alpha$ maximal ($\alpha$ is called the *inconsistency degree* of $J \cup E$).
The $\alpha$-nogood $[E \ \alpha]$ is *minimal* iff there is no $\beta$-nogood $[E' \ \beta]$ such that $E \supset E'$ and $\alpha \leq \beta$.

**Labels:** The label of a datum d noted $L(d)=\{[E_i \ \alpha_i], i \in I\}$ is the only subset of the set of environments which satisfies the following properties:

- *(weak) consistency*: $\forall \ [E_i \ \alpha_i] \in L(d)$, $J \cup E_i$ is $\beta$-inconsistent with $\beta<\alpha_i$ ($J \cup E_i$ has an inconsistency degree which is strictly less then the certainty degree obtained for d from $J \cup E_i$ (d is deduced by using a consistent sub-base of $J \cup E_i$)).

- *soundness*: $\forall \ [E_i \ \alpha_i] \in L(d)$, $[E_i \ \alpha_i]$ is an environment of d.

- *completeness*: $\forall$ E' such that d can be deduced from $J \cup E'$ with a degree $\alpha'$, then $\exists \ [E_i \ \alpha_i] \in L(d)$ such that $E' \supset E_i$ and $\alpha' \leq \alpha_i$ (all the minimal $\alpha$-environments of d are in L(d)).

- *minimality*: L(d) does not contain two environments $[E_1 \ \alpha_1]$ and $[E_2 \ \alpha_2]$ such that $E_2 \supset E_1$ and $\alpha_2 \leq \alpha_1$ (L(d) contains only the most specific $\alpha$-environments of d).

By ordering the environments in the labels on the base of their weight, a $\Pi$-ATMS can determine the set of hypotheses which allows to deduce a given datum with the greatest certainty.

**Contexts:** The context of a set of weighted hypotheses H is the set of pairs $(d, val_{[H]}(d))$, where d is a datum and $val_{[H]}(d)=\max\{\alpha$ such that d can be deduced from $J \cup H$ with a degree $\alpha\}$.

Let us refine the previous example with the weighted justification: $A \wedge B \rightarrow C \ \alpha_J$.
The label of the conclusion C will be updated by combining its previous label with the labels of the premises A and B.

The weight of the environments in this label will depend on the weights of the environments of the justification left-hand-side data and the weight of the rule conclusion as shown in figure 2.

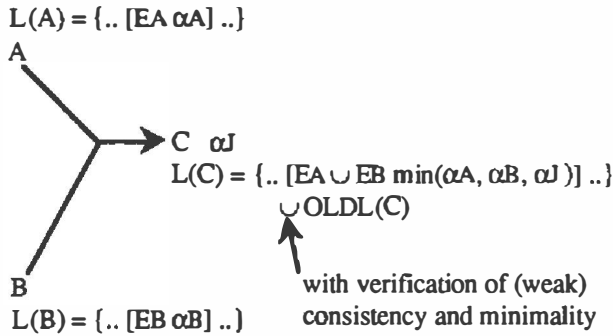

Figure 2: Propagating labels in a Π-ATMS

## 5 COUPLING A Π-ATMS AND AN INFERENCE ENGINE

ATMSs draw inferences and build multiple contexts based on initial facts, hypotheses, and justifications. Justifications could be viewed as simple rules without variables (Propositional Logic). However, most applications require more expressiveness power and rely on an OPS-like forward chaining inference engine with first order rules.

These engines encode usually a *match-select-act* cycle.

The propositions contained in the fact base may match some rules condition parts, and thus instantiate one or more rules. Instantiated rules are queued in a so called conflict set for future selection and eventual firing.

When a rule is fired, it may create new facts which will in turn instantiate new rules.

In this framework, justifications are dynamically generated and link facts created by a rule's right-hand-side to the facts which instantiated the rule.

We thus implemented a toolkit integrating an inference engine and a Π-ATMS as shown in figure 3. A similar architecture in the case of a standard ATMS is presented in [Morgue 91].

The user can create uncertain hypotheses, facts, and rules.

Facts and hypotheses are stored as Π-ATMS nodes and corresponding working memory elements are created in the inference engine and can eventually match rules conditions.

When a rule is selected and then fired, its action part does not modify directly the working memory of the inference engine (as it normally does in the standard match-select-act cycle).

Instead, new uncertain facts and hypotheses can be created and associated to new working elements, or new justifications can be installed on existing facts and hypotheses.

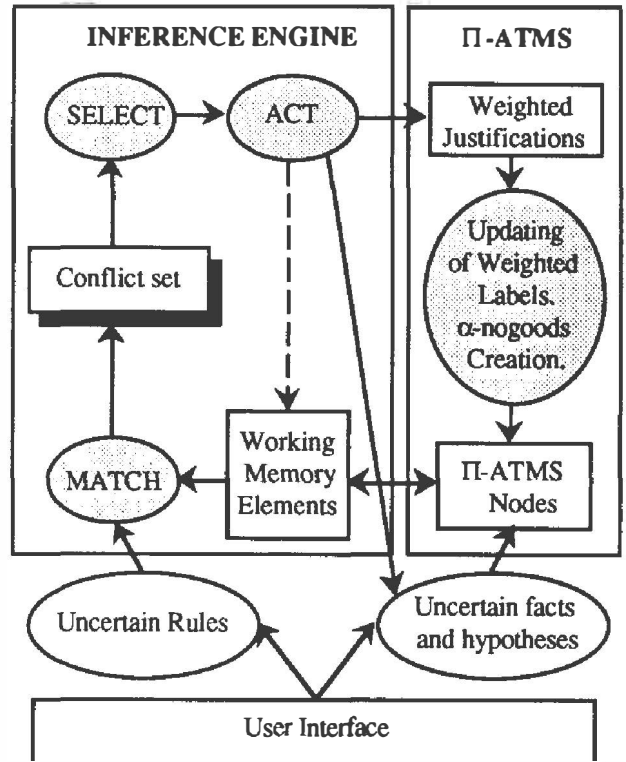

Figure 3: Coupling a Π-ATMS and an OPS-like inference engine.

The role of the inference engine is thus to produce the weighted justifications whilst the role of the Π-ATMS is to manage the uncertainty pervading the Π-ATMS nodes (facts, assumptions and justifications) by updating weighted environments in labels, handling weighted contradictions, etc.

## 6 THE AGGREGATION PHASE IN SEFIR

We are now going to describe how the proposed toolkit architecture supports the new SEFIR expert system. Units are represented as uncertain facts with the following informations:

- *id:* internal reference of the unit;
- *level:* section, company, battalion, regiment or division;
- *type:* tank, motorised rifle, etc;
- *time:* interval of time during which the unit has been observed;
- *axis:* axis on which the unit has been observed to move;
- *sub-units:* units of the lower level composing the unit itself.

The uncertainty degree of each unit is computed from its type, its level, its completeness (number and type of sub-units) and the covered spatio-temporal surface.

In SEFIR the generation of weighted hypotheses is limited by applying domain knowledge. However, a significant number of hypotheses is produced at each level in the enemy organisation hierarchy, entailing intractable ∏-ATMS computations. It is thus crucial to select only the 'best' aggregation hypotheses, for an efficient assistance in the enemy threat assessment by the operator.

We started by breaking down the global problem in smaller, more tractable ones. This was eased by the use of the Thomson-CSF proprietary XIA inference engine.
XIA allows to structure the knowledge base in many independent rule-bases (or Knowledge Sources), each of which can infer on many independent private working memories (pwm). A similar functionality is provided by the GBB-OPS5 integrated problem solving architecture [Corkill 90].

The reasoning process has been structured in order to handle the different levels of units in different private working memories.
Each level is treated separately by a specialized knowledge source with rules which formalize:

- the available knowledge on the enemy organisation at that level;
- the constraints which restrict the aggregation of units of that level into units of the higher hierarchical level.

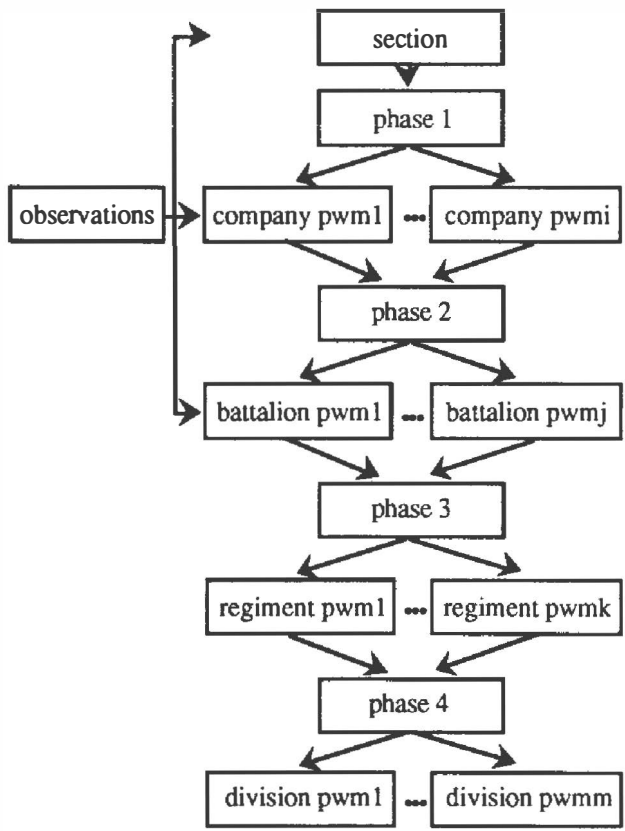

Figure 4: Aggregation in SEFIR.

The possible aggregations done at each level have been segmented in different phases:

- aggregation of sections into companies;
- aggregation of companies into battalions;
- aggregation of battalions into regiments;
- aggregation of regiments into divisions.

As shown in figure 4, at the end of each of these phases the ∏-ATMS allows to identify just the most certain among the possible combinations and to eliminate the ones which would have been most probably discarded by an expert.
This procedure is less discriminating than a real expert, but the exponential growth of possible aggregations (when the number of observations grows) is prevented.
However, by augmenting the number of selected combinations at each level a more exhaustive search and analysis of potential alternative solutions can be performed.

It is important to note that no adhoc technique is used for the selection step. The ATMS first generates all interpretations or maximal sets of non-(partial)contradictory assumptions. The certainty weights are then used to rank the interpretations and the first k, which correspond to the k most certain solutions, are selected for the next aggregation phase.
Moreover, when the operator is interested in just the best solution, the ∏-ATMS provides a much more efficient algorithm than the standard ATMS interpretations computation:

The ∏-ATMS starts with the set E of all assumptions and the set N of all nogoods.

- While N is non empty:
- it selects the most certain nogood and removes it from N;
- it then removes from E the least certain assumption involved in the selected nogood;
- it finally removes from N all nogoods involving the selected assumption.

The resulting E is the best maximal set of assumptions.

The complexity of the algorithm used for the computation of all the interpretations is upper bounded by the product of the cardinality of the nogoods whilst the complexity of the above algorithm is upper bounded by the square of the cardinality of the nogoods.

We detail in the following the aggregation mechanism of units of level n in units of level n+1, and illustrate it in the case of battalions to regiments aggregation.
The best possible aggregations of units of level n ordered by certainty are treated separately in different private working memories to produce the best possible aggregations of units of level n+1:

- For each private working memory of level n (as show in figure 5 for a battalion pwm):

- the inference engine computes, starting from the units of this private working memory and using the rules of the level n, all the hypotheses of possible *complete* units for level n+1 which are compatible with the aggregation and spatio-temporal constraints;

- the Π-ATMS computes the best k maximal consistent combinations of these hypotheses, using the fact that two hypotheses are contradictory if they correspond to two units which have at least one common sub-unit;

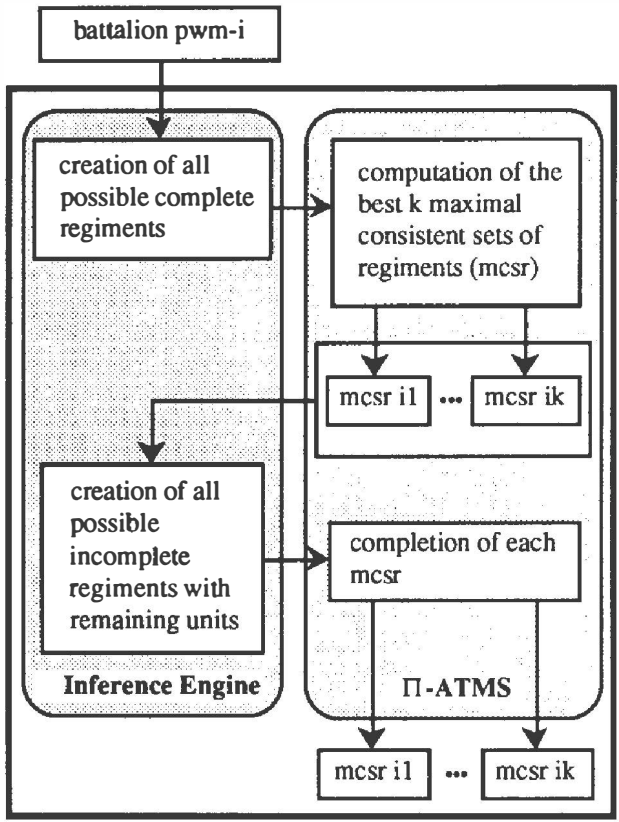

Figure 5: Computing the best k maximal consistent sets of regiments (mcsr), starting from a battalion private working memory (pwm).

- For each of these best consistent combinations:

- the inference engine computes, starting from the units of the private working memory and using the rules of the level n, all the hypotheses of possible *incomplete* units for level n+1 which are compatible with the aggregation and spatio-temporal constraints and which are not in the current consistent combination;

- the Π-ATMS completes the current consistent combination with the best newly created hypotheses to obtain the final consistent combination.

- Finally, all the best combinations produced by the different private working memories are sorted by certainty. The best k generate private working memories for level n+1: each hypothesis of the selected combination is asserted as a fact in the new working memory.

The battalions to regiments aggregation phase is illustrated in figure 6.

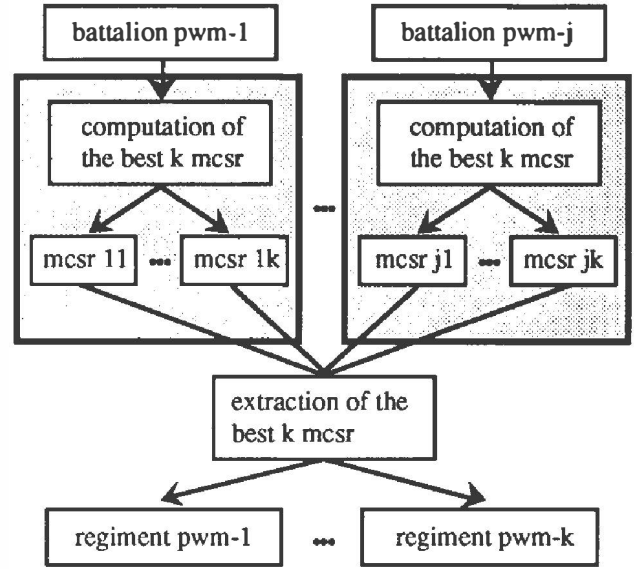

Figure 6: Aggregation of battalions into regiments (phase 3).

This cycle is repeated for each phase until the best m solution for the division level are obtained. Then the operator can evaluate the enemy threat by analysing the proposed division hypotheses.

## 7 RESULTS

In the first SEFIR prototype, the knowledge engineer had to hand code the generation and handling of multiple uncertain hypotheses.
This resulted in an unreadable knowledge base, difficult to tune and maintain, and with adhoc algorithms non reusable for similar data fusion applications.
Since the reduction of the generated uncertain hypotheses was done heuristically, the algorithm failed in some situations by eliminating solutions which were among the best possible ones (see the description of the solutions for a test scenario in the next paragraph).

The integration of a Π-ATMS and the structuring of the reasoning process were two determinant steps to achieve much better efficiency and solution quality.
The uncertain hypotheses management and their reduction is completely supported by the Π-ATMS with documented semantics, proven algorithms and logical foundations.

Moreover, the Π-ATMS is sensitive to the relative order of the weights associated to the hypotheses rather than to their effective values allowing an easier tuning of the criterias for their evaluation.

Finally, the number of solutions maintained at each phase can be defined by the operator enabling the development of all the desired alternative solutions.

The response time thus depends on the operator's choice, generating just the best solution (as explained previously) in a crisis situation is an order of magnitude faster than with the previous SEFIR prototype, while an exhaustive search can be much more costly.

## 8 A TEST SCENARIO

We detail in the following the solutions given by the first SEFIR prototype and the ones given by the possibilistic version for a test scenario in which three intelligence officers provided informations on the enemy units they observed during six hours.

The symbols used to represent the observed and aggregated units in the figures are the following:

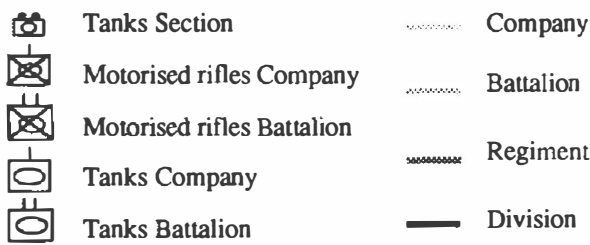

The criteria used for the evaluation of the certainty of an aggregated unit takes into account:

- the unit level and type;
- the covered spatio-temporal surface;
- the unit completeness.

We can note that this criteria doesn't take into account the completeness of the sub-units composing the aggregated unit. Thus it is not optimal.

However, since the first SEFIR prototype used this criteria, we retained it to better compare the solutions provided by the two prototypes.

The best (with respect to the criteria) three solutions given by the first SEFIR prototype are represented, respectively, in the figures 7, 8 and 9.

First of all we remark that the second solution is a subset of the first one since the grayed regiment of the figure 8 is included in the grayed regiment of the figure 7.
Thus, this second solution should not appear in the solutions proposed to the operator.

Moreover, the third solution (see figure 9) is not optimal since the aggregation of the companies noted C1, C2 and C3 in a battalion and its insertion in the grayed regiment represents a better alternative.

The presence of these solutions is due to the fact that the reduction of the generated uncertain hypotheses was done heuristically in this first prototype.

The best three solutions given by the possibilistic SEFIR prototype for the same scenario are represented, respectively, in the figures 10, 11 and 12.

The first solution is better than the corresponding one proposed by the first prototype since the regiment noted R1' in the figure 10 is more compact than the regiment noted R1 in the figure 7.
Thus the certainty of R1' is greater than the certainty of R1 (with respect to the criteria).
The second solution represents an effective alternative to the first one since the battalions composing the regiment noted R1" (see figure 11) are not the same composing the regiment noted R1' (see figure 10).
Finally, the third solution is correct since the regiment noted R2 (see figure 12) is complete. In fact R2 includes the companies noted C1, C2 and C3 in the figure 9.

As a final remark we can say that the solutions given by the possibilistic SEFIR prototype are really compatible with the selected criteria.

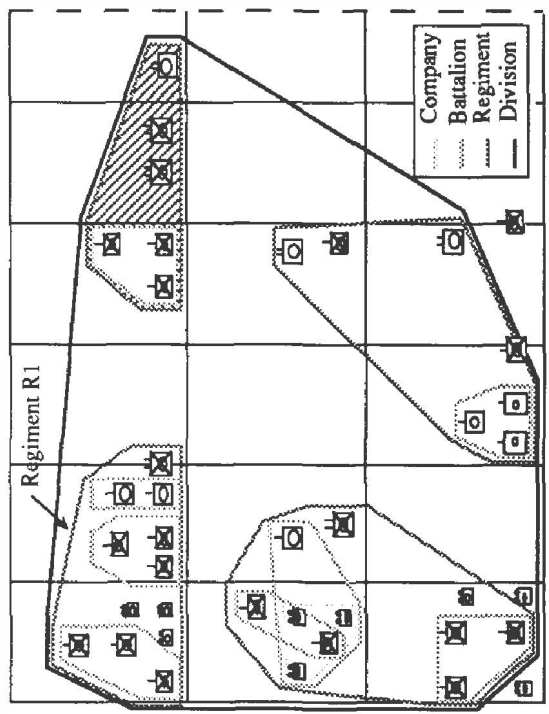

Figure 7: First solution given by the first SEFIR prototype

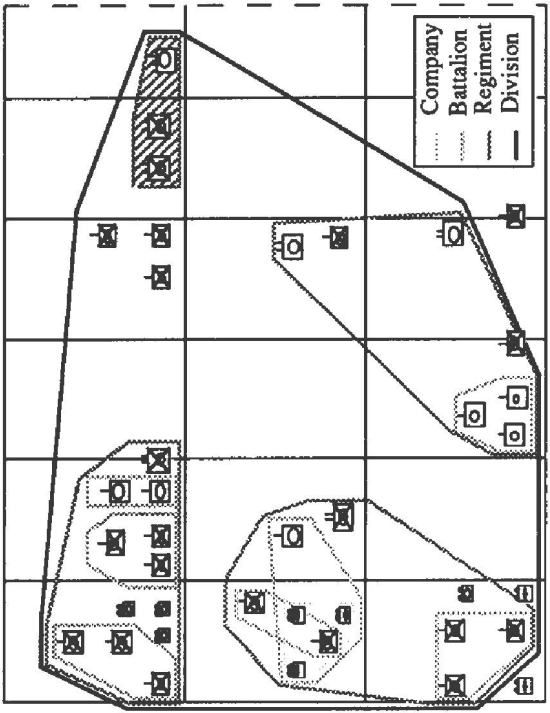

Figure 8: Second solution given by the first SEFIR prototype

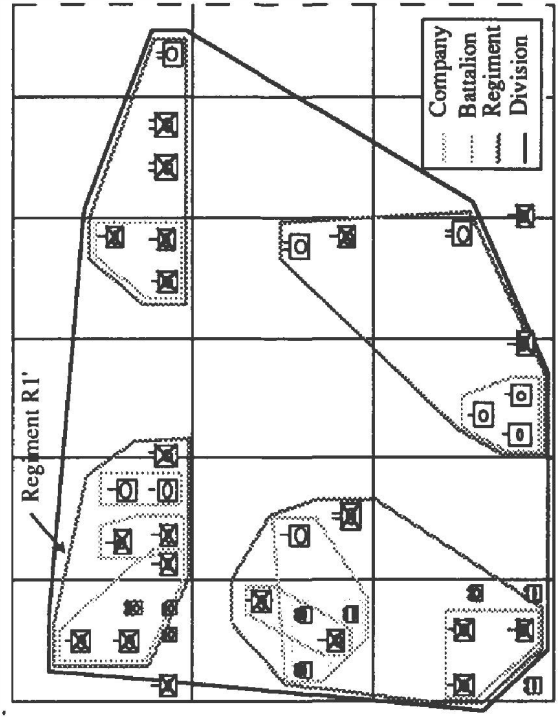

Figure 10: First solution given by the possibilistic SEFIR prototype

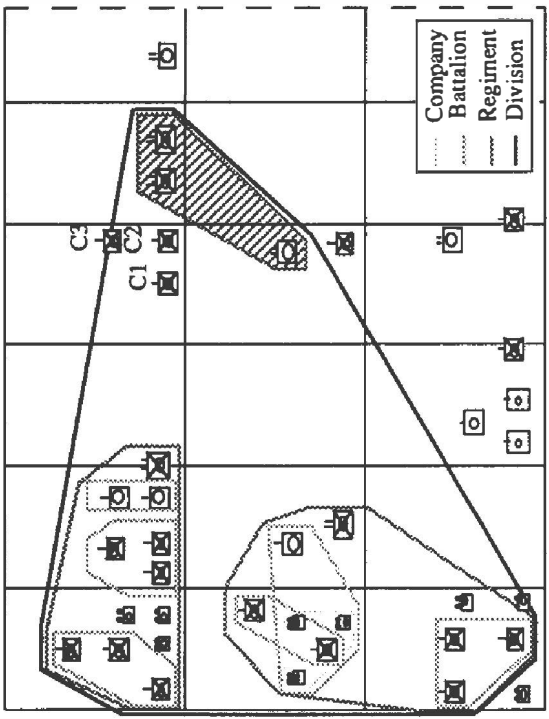

Figure 9: Third solution given by the first SEFIR prototype

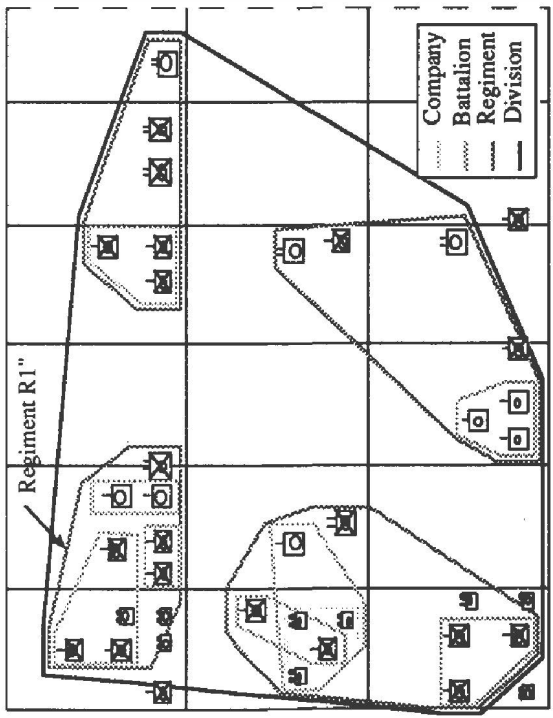

Figure 11: Second solution given by the possibilistic SEFIR prototype

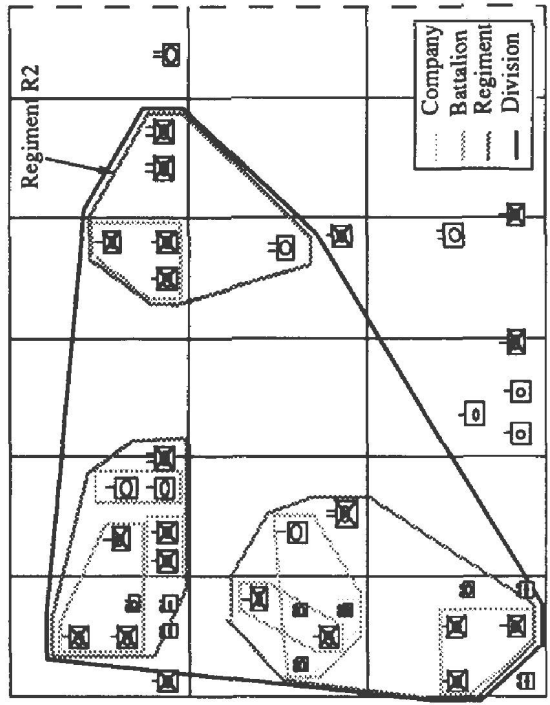

Figure 12: Third solution given by the possibilistic SEFIR prototype

## 9 CONCLUSIONS

The use of an architecture coupling a $\Pi$-ATMS and a first order inference engine enables a quicker development of more readable, efficient, maintainable and reusable solutions to the problems of uncertain data management and multiple hypotheses handling in data fusion applications.

The approach presented in this paper is well adapted to the aggregation phase of a data fusion process. On the other hand, the specific needs of the correlation phase are not covered here.

Moreover it is necessary to be able to take into account the dynamics of the informations provided to the data fusion system: timestamped data, data persistence and obsolescence, temporal constraints management.

Others issues include informations pertinence, evaluation of their (eventual) time varying certainty, etc.

Thus two main extensions to the SEFIR application are under consideration:

- the determination of plausible explanations for the observed units of intermediary hierarchical level through Abductive Reasoning.
  In the current implementation, an incoming message generates a new observed unit which is added to system deduced units of the same hierarchical level. However, searching for units of the lower level who could justify the new observation, it is sometimes possible to find an already aggregated unit that matches the observation. The system could then increase the confidence in the existing unit instead of creating a new one for the observation under consideration;

- the explicit manipulation of time for reasoning on the dates of the observations and of the system deduced units to find out eventual redundancies or inconsistent situations.

## Acknowledgements

The work described in this paper has been partially funded by the DRET (French MOD).
The inference engine component of the proposed architecture was built on top of XIA which results partly from ESPRIT project P96 and from Thomson Strategic Project in AI.
We would like to thank Pr Didier Dubois, Pr Henri Prade, Dr Jerome Lang, Mr. Salem Benferhat of IRIT (Paul Sabatier University - Toulouse - France); Mrs. Sylvie Martin dit Neuville, Mrs. Geneviève Morgue, and Mr. Bertrand Pochez of Thomson-CSF/RCC for their valuable cooperation to the development of the possibilistic version of the SEFIR expert system.